\documentclass[journal]{IEEEtran}

\usepackage{cite}
\usepackage{makecell}
\usepackage{amsmath}
\usepackage{amssymb}
\usepackage{algorithm}
\usepackage{algorithmic}
\usepackage{tikz,mathpazo}
\usepackage{url} 
\usetikzlibrary{shapes.geometric, arrows}

\begin{document}

\title{Semantic Task Planning for Service Robots in Open World}
\author{Guowei Cui, Wei Shuai and Xiaoping Chen \thanks{The authors are with the School of Computer Science and Technology, University of Science and Technology of China, Hefei, China (cuigw@mail.ustc.edu.cn; swwsag@mail.ustc.edu.cn; xpchen@ustc.edu.cn).}}

\maketitle


\begin{abstract}
	In this paper, we present a planning system based on semantic reasoning for a general-purpose service robot, which is aimed at behaving more intelligently in domains that contain incomplete information, under-specified goals, and dynamic changes. First, Two kinds of data are generated by Natural Language Processing module from the speech: (i) action frames and their relationships; (ii) the modifier used to indicate some property or characteristic of a variable in the action frame. Next, the goals of the task are generated from these action frames and modifiers. These goals are represented as AI symbols, combining world state and domain knowledge, which are used to generate plans by an Answer Set Programming solver. Finally, the actions of the plan are executed one by one, and continuous sensing grounds useful information, which make the robot to use contingent knowledge to adapt to dynamic changes and faults. For each action in the plan, the planner gets its preconditions and effects from domain knowledge, so during the execution of the task, the environmental changes, especially those conflict with the actions, not only the action being performed, but also the subsequent actions, can be detected and handled as early as possible. A series of case studies are used to evaluate the system and verify its ability to acquire knowledge through dialogue with users, solve problems with the acquired causal knowledge, and plan for complex tasks autonomously in the open world.
	
\end{abstract}

\begin{IEEEkeywords}
	General purpose service robot, task planning, open world.
\end{IEEEkeywords}

\section{Introduction}
\IEEEPARstart{I}{N} recent years, research on service robots has received increasing attention, including autonomous robots, human-robot interaction (HRI), vision, manipulation, machine learning, reasoning, and automated planning. In most scenarios, like nursing homes and offices, humans hope that robots can help them do a lot of tasks, which include taking orders and serving drinks, welcoming and guiding guests, or just cleaning up. To achieve that goal, a service robot requires human-like information processing and the underlying mechanisms for dealing with the real world, especially the ability to communicate with humans, acquire the knowledge to complete tasks and adapt to the dynamic environment.

Not surprisingly, for most users, speech is preferable to any other means of communication with a robot. The user verbally assigns a complex task to the robot that may require a set of skills. The Robot needs to perform the task, report any problems, adapt to unexpected changes, and find alternative solutions with brief knowledge about the domain. Automated planning, which makes decisions about how to act in the world, requires symbolic representations of the robot’s environment and the actions the robot is able to perform, has been widely used for task planning and control in many service robot applications. In the open world, there are two main challenges for task planning: 1) the robot's perception of the world is often incomplete, a command may refer to an object that is not in its knowledge base, lack of information will fail to generate a plan; 2) changes in the dynamic environment may not be expected by the robot, which will cause the planned action to fail.

In this paper, we address these issues by developing a semantic task planing system, which combines natural language understanding, task-oriented knowledge acquisition, and semantic-based automated task planning. For the first problem, the idea is ``close" the world, which means each object involved in the command must be known in the knowledge base. \textit{Assumption} and \textit{grounding} operations are involved to handle this. First, natural language understanding module will generate two outputs: 1) action frames and their relationships; 2) modifier used to indicate some property or characteristic of a variable in the action frame. Next, the action frame and modifiers are used to generate \textit{goals} of the task. For the object in the command that is not in the knowledge base, an assumption will be added to the knowledge base. A grounding operation will finally check this assumption is true or not when the operation is executed.

For the second problem, the environment is dynamic, robots must be able to start from incomplete information, gather useful information, and achieve the goals. In order to response the dynamic environment, \textit{continuous perception} and \textit{conflict detection} are adopted. We formalize continuous sensing in a formal representation, which is transformed into Answer Set Programming (ASP) \cite{gelfondStableModelSemantics1988} to generate plans by an ASP solver \cite{gebserClingoASPControl2014}, and the robot perform plans using the classical "plan-execute-monitor-replan" loop. The monitor checks if the change conflicts with actions in the plan, not only the action being performed, but also the subsequent actions, so the conflict can be detected and handled as early as possible. Our method features: 1) a method of confirming task type, extracting the roles of the task and the roles' constrained information; 2) assumption and grounding methodology to "close" the open-world; 3) continuous sensing and conflict detection mechanism that captures dynamical changes in the environments and triggers special processing.

This paper is organized as follows. We discuss the related work in Section \ref{section:related} and describe the overview of the system in Section \ref{section:overview}. Next, we describe the knowledge representation and domain formulation in Section \ref{section:knowledge} and natural language understanding in Section \ref{section:hri}. Then in Section \ref{section:plan}, where the implemented ``plan-execute-monitor-replan" techniques are described. Experimental results and evaluations are presented in Sections \ref{section:experiments}.

\section{Related Work}\label{section:related}

Research in knowledge representation (KR) and logical reasoning has provided sophisticated algorithms \cite{ghallabAutomatedPlanningTheory2004,gelfondKnowledgeRepresentationReasoning2014,sridharanUsingKnowledgeRepresentation2016}, which have been used on service robots for supporting task planning and execution. The Kejia robot \cite{chenIntelligentTechniquesRobot2015} represents domain knowledge learned through natural language processing and leverages a symbolic planner for problem-solving and planning to provide high-level functions \cite{chenDevelopingHighlevelCognitive2010,chenOpenKnowledgeEnabling2013}. The system has been extended to acquire task-oriented knowledge by interacting with the user and sensing the environment \cite{chenPlanningTaskorientedKnowledge2016}. By interacting and sensing and grounding useful sensing information, the robot can work with incomplete information and unexpected changes. Savage et al. \cite{savageSemanticReasoningService2019} use a conceptual-dependency \cite{schankConceptualDependencyParser1969} interpreter extracts semantic role structures from the input sentence and planning with the open-source expert system CLIPS \cite{culbertCLIPSReferenceManual1988}. Puigbo et al. \cite{puigboUsingCognitiveArchitecture2015} adopt Soar cognitive architecture \cite{lairdSoarCognitiveArchitecture2012} to support understanding and executing human-specified commands. Similar to these works, our system combines a KR system and an ASP planner for high-level planning.

Planning approaches that work in open-world scenarios often need to find a way to close the world. Using Open World Quantified Goal (OWQGs), \cite{talamadupulaIntegratingClosedWorld2010,talamadupulaPlanningHumanrobotTeaming2010} can bias the planner’s view of the search space towards finding plans that achieve additional reward in an open world. To address incomplete information in the open world, methods of planning with HRI and sensing actions are developed. Petric et al. utilize a collection of databases, each representing a different kind of knowledge \cite{petrickKnowledgebasedApproachPlanning2002}. Some methods \cite{puigboUsingCognitiveArchitecture2015,thomasonLearningInterpretNatural2015} collects information from user during natural language processing (NLP). representing HRI actions using planning actions \cite{sanelliShorttermHumanrobotInteraction2017}. Some works collects task-oriented information by combining HRI with planning \cite{brennerContinualPlanningActing2009,brennerMediatingQualitativeQuantitative2007,brennerSituationAwareInterpretationPlanning2007,chenPlanningTaskorientedKnowledge2016}. Some work focus on using open source knowledge to handle the incomplete information \cite{luUnderstandingUserInstructions2016,chenOpenKnowledgeEnabling2013}. Planning with sensing actions has been investigated under different semantics and specific planning algorithms \cite{petrickKnowledgebasedApproachPlanning2002,hoffmannContingentPlanningHeuristic2005, sonPlanningSensingActions2002}. Mininger and Laird \cite{miningerInteractivelyLearningStrategies2016} use a Soar-based interactive task-learning system to learn strategies to handle references to unseen objects. The approach defines a ``find” subtask with a special postcondition so the system can succeed in planning for tasks requiring direct interaction with unseen objects. In \cite{cashmoreRosplanPlanningRobot2015}, sensor information is used to update an ontology that is queried in each planning loop to populate a PDDL \cite{foxPDDL2ExtensionPDDL2003} problem file. Petrick et al. \cite{petrickExtendingKnowledgeLevelContingent2014} extensions to the knowledge-level PKS (Planning with Knowledge and Sensing) planner to improve the applicability of robot planning involving incomplete knowledge. Hanheide et al. \cite{hanheideRobotTaskPlanning2017} extends an active visual object search method \cite{aydemirActiveVisualObject2013} to explain failures by planning over explicitly modeled additional action effects and assumptive actions. Jiang et al. \cite{jiangOpenWorldReasoningService2019} provide an open-world task planning approach for service robots by forming hypotheses implied by commands of operators. Similar to \cite{hanheideRobotTaskPlanning2017,jiangOpenWorldReasoningService2019}, assumptions and grounding actions are adopted in this paper, but only when there are not enough objects to meet the user's instructions, the corresponding assumptions will be generated. A soft set is used to track the objects of each type of condition.

Another research area that is related to this work focuses on planning with dynamic change and uncertainty. The work of Zhang et al. uses commonsense reasoning to dynamically construct (PO)MDPs for adaptive robot planning \cite{zhangDynamicallyConstructedPO2017,zhangArchitectureKnowledgeRepresentation2014}. Using theory of intensions \cite{blountTheoryIntentionsIntelligent2015}, which is based on scenario division, \cite{srivastavaTractabilityPlanningLoops2015} generates plan with loops to accommodate unknown information at planning time. In this paper, a classic ``plan-execute-monitor-replan'' is used to handle unexpected changes in the open world.

\section{Framework Overview}\label{section:overview}

An architecture that is briefly introduced in this section. A central problem in robotic architectures is how to design a system to acquire a stable representation of the world suitable for planning and reasoning.

In general, a service robot should be able to perform basic functions:
\begin{itemize}
    \item Self-localization, autonomous mapping, and navigation.
    \item Object detection, recognition, picking, and placement.
    \item People detection, recognition, and tracking.
    \item Speech recognition and Natural Language Understanding.
\end{itemize}

A long-standing challenge for robotics is how to act in the face of uncertain and incomplete information, and to handle task failure intelligently. To deal with these challenges, we propose a framework for service robots to behave intelligently in domains that contain incomplete information, under specified goals and dynamic change. The framework is composed of five layers: Input, Knowledge Management, Planning, Execution, and Monitor.

In this section, we will describe the most relevant modules categorized by layer.

\subsection{Input Layer}
This layer involves the modules that provide basic facilities for sensing its environment and communication with other agents, mainly includes two modules: perception and human-robot interaction.

\subsubsection{Perception}
This module aims to sense the environment, which has several sub-modules as follows.
\begin{itemize}
    \item Self-localization and autonomous mapping.
    \item Object detection and recognition.
    \item People detection and recognition.
\end{itemize}

This module generates a set of beliefs about the possible states of the environment. Beliefs are based on the symbolic representation of the sensorial information coming from internal and external sensors. These beliefs are transferred to Knowledge Management and used to update the state of the world.

\subsubsection{Human Robot Interface}
This module contains two parts: Speech Recognition and Natural Language Understanding. HRI provides the interface for communication between users and the robot.

Speech Recognition uses the Speech Application Programming Interface (SAPI) which developed by iFlytek The speech will be processed in the NLU module as explained in Section \ref{section:hri}. Based on the user's instructions, NLU generates a set of assumptions about the possible states of the environment and goals that represent the user's intents. These assumptions are based on the symbolic representation of the information coming from users. These assumptions are transmitted to the Knowledge Management, and goals are transmitted to Monitor to trigger a plan.

\subsection{Knowledge Management Layer}
This layer involves all modules that store and provide access to the robot’s knowledge. Such knowledge, which is symbolic, includes the structuring of the informational state of the world, goals, and domain knowledge.

For high-level reasoning, a rule-based system is used. The facts and rules are written in ASP \cite{gelfondStableModelSemantics1988} format and represent the robot’s knowledge as explained in detail in Section \ref{section:knowledge}.

\subsection{Planning Layer}
This layer is responsible for generating plans at a high level of abstraction and performing global reasoning.

Beliefs generated by the perception module and assumptions generated by the HRI are transferred to Knowledge Management as states. Together with domain knowledge, they are used to trigger the Action Planner, which will generate a sequence of actions to achieve the desired goals.

\subsection{Monitor Layer}
This layer dispatches the generated plan to the execution layer, monitors the execution and changes in the open world. By “plan-execute-monitor-replan”, if something unexpected happens while executing a plan, the monitor will interrupt the execution and trigger the generation of a new plan.

\subsection{Execution Layer}
This layer controls the robot to execute the generated plans. Each step of the plan is an atomic function that solves a specific problem. These functions should be simple, reusable, and easy to implement with a state machine.


\section{Natural Language Understanding}\label{section:hri}

This section describes the NLU technology employed by this work. The NLU is in charge of translating the speech from users into a symbolic representation that can be used by the action planner.

For each sentence, NLU finds the main event. After finding the main event in such a sentence, it must determine the roles played by the elements of the sentence, and the conditions under which the event takes place. The verb in a sentence usually is used to find a structure of the event, that is composed of participants, objects, actions, and relationships between event elements, these relations can be temporal or spatial. For example, in ``\textit{Robot, find Mary, and bring her an apple}", \textit{Robot} is the actor, \textit{her} is the recipient, \textit{apple} is the object, and \textit{an} represents the number of the \textit{apple}. According to the context, the NLU module should be able to figure out who \textit{her} is referring to.

For each input sentence, the NLU module works in three steps: (1) Parsing, in which Stanford Parser \cite{kleinFastExactInference2002} parses the sentences and outputs grammatical relations as typed dependencies; (2) Semantic analysis, in which typed dependencies are used to generate action frames; (3) Goals generation, in which action frames are translated into the logic predicates that can be recognized by an ASP solver.

\subsection{Parsing}
The input of the NLU module from the human-robot dialog is a string of words that is regarded as a sentence. This sentence is parsed by the Stanford parser, which works out the grammatical structure of sentences, can offer two kinds of information: a grammar tree with UPenn tagging style, and a set of typed dependencies with Universal Dependencies style or Stanford Dependencies style. These typed dependencies are otherwise known grammatical relations.

In our system, we use universal dependencies. The idea of universal dependencies is to propose a set of universal grammatical relations that can be used with relative fidelity to capture any dependency relation between words in any language. There are 40 universal relations, here's a brief introduction to part relationships that play an important role in semantic analysis.

The core dependencies play the most important role to get semantic elements of an action or event. We mainly consider three core dependencies: 
\begin{itemize}
    \item \textit{nsubj}: nominal subject. The governor of this relation is a verb in most cases, and it may be headed by a noun, or it may be a pronoun or relative pronoun.
    \item \textit{dobj}: direct object. Typically, the direct object of a verb is the noun phrase that denotes the entity acted upon or which changes state or motion.
    \item \textit{iobj}: indirect object. In many cases, the indirect object of a verb is the recipient of ditransitive verbs of exchange.
\end{itemize}

Modifier word is also an important type of dependency, we consider \textit{amod}, \textit{nummod}, \textit{det}, \textit{neg}, \textit{nmod} in our system.
\begin{itemize}
    \item \textit{amod}: adjectival modifier. An adjectival modifier of a noun is an adjectival phrase that serves to modify the meaning of the noun.
    \item \textit{nummod}: numeric modifier. A numeric modifier of a noun is any number phrase that serves to modify the meaning of the noun with a quantity.
    \item \textit{det}: determiner. The relation determiner (det) holds between a nominal head and its determiner. Determiners are words that modify nouns or noun phrases and express the reference of the noun phrase in context. That is, a determiner may indicate whether the noun is referring to a definite(words like many, few, several) or indefinite(words like much, little) element of a class, to an element belonging(words like your, his, its, our) to a specified person or thing, to a particular number or quantity(like words any, all), etc.
    \item \textit{neg}: negation modifier. The negation modifier is the relation between a negation word and the word it modifies.
    \item \textit{nmod}: nominal modifier. It is a noun (or noun phrase) functioning as a non-core (oblique) argument or adjunct. This means that it functionally corresponds to an adverbial when it attaches to a verb, adjective, or adverb. But when attaching to a noun, it corresponds to an attribute, or genitive complement (the terms are less standardized here).
\end{itemize}

\subsection{Semantic Analysis}\label{section:sa}
To get the semantic representation, which is a set of semantic elements, typed dependencies are required, and sometimes the syntactic categories of words and phrases are required, too. These typed dependencies are used to generate action frames and modifiers.

\subsubsection{Action Frame}\label{section:frame}
A semantic role refers to a noun phrase that fulfills a specific purpose for the action or state that describes the main verb of a statement. The complete description of the event can be modeled as a function with parameters that correspond to semantic roles in an event that describes the verb, such as actor, object, start and destination place. An action frame is generated from typed dependencies, the frame contains five elements: \textit{action(Actor, Action, Object, Source, Goal)}.

\begin{itemize}
\item \textit{Actor}: The entity that performs the \textit{Action}, the \textit{Actor} is an agent that usually is a person or a robot.
\item \textit{Action}: Performed by the \textit{Actor}, done to an \textit{Object}. Each action primitive represents several verbs with a similar meaning. For instance, give, bring, and take have the same representation(the transfer of an object from one location to another).
\item \textit{Object}: The entity the \textit{Action} is performed on. It should be noted that the \textit{Object} can also be a person or robot. For instance, from sentence \textit{``bring James to the office"}, an action frame \textit{action(NIL, bring, James, NIL, office)} is generated, where \textit{NIL} represents an empty slot that needs to be filled according to the content and domain knowledge.
\item \textit{Source}: The initial location of \textit{Object} when the \textit{Action} starts.
\item \textit{Goal}: The final location of \textit{Object} when the \textit{Action} stops.
\end{itemize}

Usually, from the core dependencies, such as \textit{nsubj} and \textit{dobj}, an action frame's \textit{Actor}, \textit{Action}, \textit{Object} can be identified. The slots \textit{Source} and \textit{Goal}, both of them are always associated with prepositional phrase \textit{PP} and dependency \textit{nmod}.

For instance, with ``take this book from the table to the bookshelf" as an input, Stanford Parser outputs the following result.\\
\textbf{Parser tree:}
\texttt{{\\
(ROOT\\
\hspace*{10pt}(S\\
\hspace*{20pt}(VP (VB take)\\
\hspace*{30pt}(NP (DT this) (NN book))\\
\hspace*{30pt}(PP (IN from)\\
\hspace*{40pt}(NP\\
\hspace*{50pt}(NP (DT the) (NN table))\\
\hspace*{50pt}(PP (TO to)\\
\hspace*{60pt}(NP (DT the)\\
\hspace*{70pt}(NN bookshelf))))))))}}
\\
\textbf{Typed dependencies:}
\texttt{{\\
root(ROOT-0, take-1)\\
det(book-3, this-2)\\
dobj(take-1, book-3)\\
case(table-6, from-4)\\
det(table-6, the-5)\\
nmod:from(take-1, table-6)\\
case(bookshelf-9, to-7)\\
det(bookshelf-9, the-8)\\
nmod:to(table-6, bookshelf-9)
}}

The tag \textit{VB} is used to identify the \textit{Action}, and the dependency \textit{dobj} is the core relation used to determine the \textit{Object}. From \textit{nmod:from} and \textit{nmod:to}, the \textit{Source} and \textit{Goal} location of the action frame are extracted. No \textit{Actors} are found from the parsing result, obviously, the \textit{Actor} can only be identified by content, and that will be the person or robot who talks to the person who said this sentence.

\subsubsection{Modifier}
A modifier is a word, phrase or clause that modifies other elements of a sentence. Nouns, adjectives, adjective clauses and participles can be used as modifiers of nouns or pronouns; A quantifier word is used in conjunction with a noun representing a countable or measurable object or with a number, often used to indicate a category. To be more intuitive, there are some examples in table \ref{tab:mod}.

\begin{table*}[ht]
    \centering
    \caption{Some modifier examples.}
    \label{tab:mod}
    \begin{tabular}
        {p{0.12\textwidth}p{0.12\textwidth}p{0.2\textwidth}p{0.28\textwidth}}  
        \hline
        \makecell*[lc]{\textbf{Phase}} & \makecell*[lc]{\textbf{POS}} & \makecell*[lc]{\textbf{Dependency}} & \makecell*[lc]{\textbf{Description}} \\
        \hline
        \makecell*[lc]{big cup} & \makecell*[lc]{JJ big \\ NN cup} & \makecell*[lc]{amod(cup, big)} & \makecell*[lc]{The size of the cup, which is big size.}\\
        \hline
        \makecell*[lc]{two apples} & \makecell*[lc]{CD two \\ NNS apples} & \makecell*[lc]{nummod(apples, two)} & \makecell*[lc]{The number of the apples.}\\
        \hline
        \makecell*[lc]{no apples} & \makecell*[lc]{DT no \\ NNS apples} & \makecell*[lc]{neg(apples, no)} & \makecell*[lc]{Zero.}\\
        \hline
        \makecell*[lc]{my book} & \makecell*[lc]{PRP\$ my \\ NN book} & \makecell*[lc]{nmod:poss(book, my)} & \makecell*[lc]{Possession, I own this book.} \\
        \hline
    \end{tabular}
\end{table*}

An \textit{Modifier}, indicates some attribution of an object with some value. For example, dependency \textit{nummod(apples, two)} is represented as \textit{number(apple, 2)}. The word \textit{number} means the attribute, the number \textit{2} is the value, and \textit{apple} is the object. These modifiers provide conditionality for the elements in the action frame.

\subsubsection{Pronoun}
Pronouns are often used, so an important task is to figure out the noun substituted by the pronoun. It's necessary, or the robot has to ask who or what the pronoun refers to. Our system performs a primitive type of deduction according to the principle of closest matching. In every sentence, the noun, such as actor, object or location recognized is saved. When a pronoun appears in a later part or a new sentence, the closest matching in the saved nouns is used to replace the pronoun. Matching is based on the possible meaning of the pronoun itself and the restrictions in the sentence, such as the action acting on the pronoun. For instance, when the user says \textit{``grasp a cup, go to the living room, and give it to Mary"}, \textit{it} will be recognized as \textit{cup} rather than \textit{living room} due to the room is immovable.

\subsection{Goals Generation}\label{section:gen_goal}
In our system, the ultimate goal of NLU is to generate goals that should be achieved by executing the plan solved by the solver according to the current world state. The following table lists some actions and corresponding goals.

\begin{table}[ht]
    \centering
    \caption{Action frame and corresponding goal.}
    \label{tab:goal}
    \begin{tabular}
        {p{0.17\textwidth}p{0.08\textwidth}p{0.1\textwidth}}  
        \hline
        \makecell*[lc]{\textbf{Action Frame}} & \makecell*[lc]{\textbf{Goal}} & \makecell*[lc]{\textbf{Assumption}} \\
        \hline
        \makecell*[lc]{$ action(A,get,O,F,T) $} & \makecell*[lc]{$ in(O,A) $} & \makecell*[lc]{$ in(O,F) $} \\
        \hline
        \makecell*[lc]{$ action(A,give,O,F,T) $} & \makecell*[lc]{$ in(O,T) $} & \makecell*[lc]{$ \varnothing $} \\
        \hline
        \makecell*[lc]{$ action(A,move,O,F,T) $} & \makecell*[lc]{$ in(A,T) $} & \makecell*[lc]{$ in(A,F) $} \\
        \hline
        \makecell*[lc]{$ action(A,find,O,F,T) $} & \makecell*[lc]{$ in(O,F) $} & \makecell*[lc]{$ in(O,F) $} \\
        \hline
    \end{tabular}
\end{table}

The predicate $ in $ can be transformed into the required predicate according to the category of its parameters. For example, $ in(O,A) $ can converted into $ isHeld(O,A) $, which indicates $ O $ is held by $ A $, when $ O $ is an graspable object and $ A $ is a person or robot. The assumption is information derived from the user's instructions, but has not yet been confirmed by the robot. Each assumption needs to be identified by the $ find $ operator.

In the user's instructions, the object or place may not be unique, and this generic object or place constraint needs to be added to the goal representation. Besides, the objects involved in the command may have additional constraints, which are usually represented by modifiers and need to be added to the target representation. For instance, \textit{``give Mary two apples"}, its corresponding action frame is $ action(robot,give,apple,None,Mary) $ and corresponding modifier is $ number(apple,2) $, its goal in a clingo program is shown bellow.\footnote{The encodings follow the style of the examples in the CLINGO guide: https://github.com/potassco/guide}
\[ 2 \: \{ \: in(X,mary) : apple(X) \: \} \: 2. \]

\section{Knowledge Representation}\label{section:knowledge}

Knowledge in this work represents type hierarchy, domain objects, states and causal laws. The causal laws include effects of physical actions, HRI actions and
sensing actions on fluents that represent world state. We particularly focus on three kinds of knowledge. (i) \textbf{Domain knowledge}, including causal laws that formalize the transition system and knowledge of entities(such as objects, furniture, rooms, people, etc.). (ii) \textbf{Control knowledge}. When the robot is facing a dining table and about to pick up \textit{Pepsi}, it will measure the distance of the object to determine if it should move closer, adjust its gripper, or simply fetch. (iii) \textbf{Contingent knowledge}. Throughout performing the task, the robot should continuously observe the environment, gather useful information, enrich its knowledge and adapt to the change.

Answer Set Programming (ASP) \cite{gelfondStableModelSemantics1988} is adopted as knowledge representation and reasoning tool. It is based on the stable model (answer set) semantics of logic programming. When ASP is used for planning, an action model is not divided into precondition and effect like PDDL \cite{foxPDDL2ExtensionPDDL2003}. Our system needs to check preconditions and effects before or after performing each step in a plan, so we code the action model according to the precondition and effect, then convert it into an ASP program.

\subsection{Domain Knowledge}\label{section:dn}
Domain knowledge consists of two kinds of knowledge:
\begin{itemize}
    \item The information(type, position, etc.) related to objects(humans, objects, locations, rooms, and robots).
    \item Causal laws that formalize the transition system.
\end{itemize}

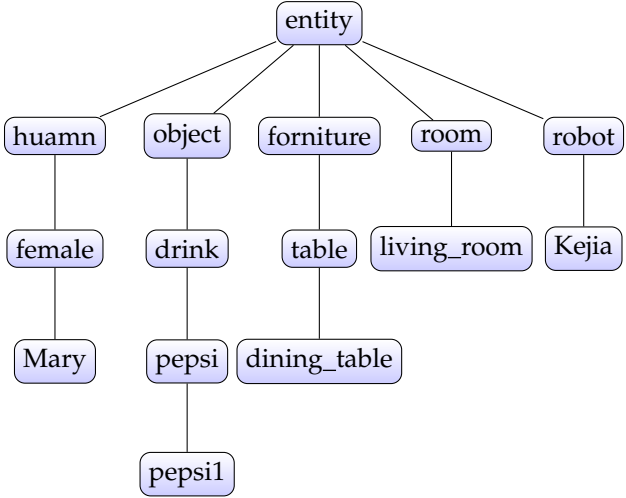
\begin{figure}[ht]
    \begin{tikzpicture}
    [sibling distance=5em,
    every node/.style = {shape=rectangle, rounded corners,
        draw, align=center,
        top color=white, bottom color=blue!20}]
    \node {entity}
    child { node {huamn} 
        child {node {female}
            child {node {Mary}}}}
    child { node {object}
        child {node {drink}
            child {node {pepsi}
                child {node {pepsi1}}}}}
    child { node {forniture}
        child {node {table}
            child {node {dining\_table}}}}
    child { node {room}
        child {node {living\_room}} }
    child { node {robot}
        child {node {Kejia}}};
    \end{tikzpicture}
    \centering
    \caption{Illustrate the type hierarchy. There are five top types in our system: robot, human, furniture, room, and object.}
    \label{fig:hierarchy}
\end{figure}

\noindent\textbf{Types}. We use a rigid (time-independent) fluent to denote type and object membership. For instance, we denote \textit{pepsi1} is a member of type \textit{pepsi} by pepsi(pepsi1). Type hierarchy and relations are formalized using static laws such as \textit{obj(X)} if \textit{pepsi(X)}. In our system, there are five top types, and each type has some sub-types. As Fig. \ref{fig:hierarchy} shows, the leaf nodes are memberships, and internal nodes are types of entities. The parent node of \textit{pspsi1} is pepsi, and \textit{pepsi}'s parent node is \textit{drink}. Different types of entities have various properties, each of them corresponds to a predicate, corresponding and different information. These properties are used for object filtering.

\noindent\textbf{States}. Non-rigid fluents, each of them means a particular situation that the object is in: the location of objects, the position relation between objects, the gripper is empty or not, the object is on a table or not, a container or door is open or closed, the robot's position, etc.
\\
\textbf{Static laws}. The transformation of fluents which are independent of actions and time. Static rules can describe category relations, inherent properties, etc. For instance, \textit{``pepsi is a drink"}, from \textit{pepsi(pspsi1)}, \textit{drink(pepsi1)} can be generated by a static causal law:
\[ drink(A) :- \: pepsi(A). \]
\noindent\textbf{Dynamic laws}. The rules of describing the changing of fluents over time or actions, play an core role in action planning. These rules fall into two categories:
\begin{enumerate}
    \item The changing of states over time. In our system, without the influence of actions, states will remain unchanged.
    \item The transition under an operation.
\end{enumerate}
The transition with an action, have the following basic components:
\begin{enumerate}
    \item \textit{Action}, an operation that the robot can perform to interact with the environment.
    \item \textit{Preconditions}, a set of states, that need to be met before or during an action.
    \item \textit{Effects}, a set of states that can be achieved at the end of an action.
\end{enumerate}

The following is a subset of actions in our system.

\begin{enumerate}
    \item \textbf{findPerson(H,L)}: The robot searches the target person \textit{H} at the location \textit{L}.
    \item \textbf{findObj(O,L)}: The robot activates object recognition to find a target object \textit{O} from the location \textit{L}.
    \item \textbf{moveTo(L,R)}: The robot navigates to the place \textit{L} in the room \textit{R}.
    \item \textbf{moveIn(R2,R1,D)}: The robot navigates to the room \textit{R2} through the door \textit{D} from the room \textit{R1}.
    \item \textbf{pickup(O,L,G)}: The robot pick up the object \textit{O} from the location \textit{L} that it will carry in its actuator \textit{G}.
    \item \textbf{putdown(O,G,L)}: The robot releases the object \textit{O} that it carries in its actuator \textit{G} in the specified place \textit{L}.
    \item \textbf{give(O,G,H)}: The robot hands the person \textit{H} the object \textit{O} in its hand \textit{G}.
\end{enumerate}

Table \ref{tab:action} lists their preconditions and effects. The action models are also generated in our system, and each change of the world is detected to see if it conflicts with the actions' preconditions in the plan. How to deal with this conflict will be discussed later.

\begin{table*}[ht]
    \centering
    \caption{Action, its preconditions and effects in our system.}
    \label{tab:action}
    \begin{tabular}{p{0.25\textwidth}p{0.25\textwidth}p{0.3\textwidth}}
        \hline
        \makecell*[lt]{\textbf{Action}} & \makecell*[lt]{\textbf{Preconditions}} & \makecell*[lt]{\textbf{Effects}} \\
        \hline
        \makecell*[lt]{$ moveTo(L,R) $} & \makecell*[lt]{$inRoom(robot,R)$ \\ $inRoom(L,R)$ \\ $not \  isNear(robot,L)$} & \makecell*[lt]{$isNear(robot,L)$ \\ $\neg{isNear(robot,L1)} \  if \  isNear(robot,L1)$} \\
        \hline
        \makecell*[lt]{$ moveIn(R2,R1,D) $} & \makecell*[lt]{$ door(D,R1,R2) $ \\ $ open(D) $ \\ $ inRoom(robot,R1) $} & \makecell*[lt]{$ inRoom(robot,R2) $ \\  $ \neg{inRoom(robot,R1)} $ \\ $ \neg{~isNear(robot,L)} \ if \ isNear(robot,L) $} \\
        \hline
        \makecell*[lt]{$pickup(O,L,G)$} & \makecell*[lt]{$empty(G)$ \\ $ graspable(O) $ \\ $ isNear(robot,L) $ \\ $ isPlaced(O,L) $} & \makecell*[lt]{$ isHeld(O,robot) $ \\ $ inHand(O,G) $ \\ $ \neg{empty(G)} $ \\ $ \neg{isPlaced(O,L)} $ \\ $ \neg{inRoom(O,R)} \ if \ inRoom(robot,R) $} \\
        \hline
        \makecell*[lt]{$ putdown(O,L,G) $} & \makecell*[lt]{$ inHand(O,G) $ \\ $ isNear(robot,L) $ \\ $ isPlacement(L,true) $} & \makecell*[lt]{$ isPlaced(O,L) $ \\ $ empty(G) $ \\ $ \neg{inHand(O,G)} $ \\ $ \neg{isHeld(O,robot)} $ \\ $ inRoom(O,R) \ if \ inRoom(robot,R) $} \\
        \hline
        \makecell*[lt]{$ findPerson(H,L) $} & \makecell*[lt]{$ assume(isNear(H,L)) $ \\ $ human(H) $ \\ $ isNear(robot,L) $} & \makecell*[lt]{$ isNear(H,L) $ \\ $ \neg{assume(isNear(H,L))} $ \\ $ inRoom(H,R) \ if \ inRoom(robot,R) $} \\
        \hline
        \makecell*[lt]{$ findObj(O,L) $} & \makecell*[lt]{$ assume(isPlaced(O,L)) $ \\ $ object(O) $ \\ $ isNear(robot,L) $} & \makecell*[lt]{$ isPlaced(O,L) $ \\ $ \neg{assume(isPlaced(O,L))} $ \\ $ inRoom(O,R) \ if \ inRoom(robot,R) $} \\
        \hline
        \makecell*[lt]{$ give(O,H,G) $} & \makecell*[lt]{$ inHand(O,G) $ \\ $ isNear(robot,H) $ \\ $ human(H) $} & \makecell*[lt]{$ isHeld(O,H) $ \\ $ empty(G) $ \\ $ \neg{inHand(O,G)} $ \\ $ \neg{isHeld(O,robot)} $} \\
        \hline
    \end{tabular}
\end{table*}

\subsection{Control Knowledge}
In our system, control knowledge is oriented to atomic operations in task planning. Each atomic operation is implemented using one or more state machines. The goal of this knowledge is to accomplish tasks more efficiently. These knowledge are usually some control parameters, which are applied to the state machine to complete the atomic operations.

For instance, when the robot is performing a pick-up task, for example, grabbing a bottle of iced black tea, it will measure the distance of the object to determine if it should move closer, adjust its gripper, or just simply fetch. The robot needs to figure out that the distance of objects may affect its manipulation strategy, figure out grabbing which part and using which posture to grab to get a greater success rate. For a recognition task, like finding an apple, 
the sequence of searching places can be added to the domain language for task planning, but the angle adjustment of the camera and the distance between the object and the robot to complete the identification task more efficiently, belong to the control knowledge, which is integrated into the visual control system.

\subsection{Contingent Knowledge}
In the process of performing tasks, robots should constantly observe the environment, collect useful information, enrich knowledge and adapt to changes. This is particularly important because objects in domestic environment are constantly changing, and the information provided by human can be fuzzy or wrong. Therefore, robots must start from a local, incomplete and unreliable domain representation to generate plans to collect more information to achieve their goals.

Continuous sensing is a mechanism that updates the current state when the perception module finds new information. It allows the robot to reduce the uncertainty of the domain while executing the actions. Therefore, the robot has stronger adaptability and robustness to changing fields and unreliable actions.

The information discovered by sensing are encoded as \textbf{States} mentioned in \ref{section:dn}. There are two types of knowledge effects/states the robot’s actions can have: \textit{belief} (I believe X because I saw it) and \textit{assumption} (I’ll assume X to be true) \cite{hanheideRobotTaskPlanning2017}. Assumptions are derived from the user's speech or historical observations, and the transformation from \textit{assumption} to \textit{belief} is achieved through continuous sensing. In planning, realizer actions, such as \textit{findPerson} and \textit{findObj}, are used to complete this transformation.


\section{Planning, Execution and Monitor}\label{section:plan}

The main control loop for plan generation and execution follows the traditional \textit{planning-execution-monitor-replan} loop. Symbol grounding for sensing actions is handled the same way as in continuous observation. The execution result is compared with expected state to determine if replan is needed or goal is achieved.

\subsection{Planning}
A robot task planning problem is defined by the tuple $ (S0, G, P) $, where $ S_0 $ is the initial state, $ G $ is the goal condition, $ P $ is a plan consisting of a set of action sequences $ P = \langle a_1, . . . , a_n \rangle $. The states and actions for planning are described in Section \ref{section:dn}.

\subsubsection{Goal Condition}
The goal $G$ for planning is generated by parsing a spoken command, how to generate goal condition from a command is described in Section \ref{section:hri}.

\subsubsection{Initial State}
The initial state $ S_0 $ for planning is generated as follows: (i) fluents that belong to definite world state are initialized based on robot’s sensor inputs, aka \textit{beliefs}; (ii) fluents that generated by NLU module through talking to people or experiences, aka \textit{assumptions}; (iii) fluents that belong to belief state are initialized as negated literals, denoting that the robot does not know anything about them. All these fluents are stored in the database. When goal of the task achieves, some of fluents is extracted from the database to form the initial state. Only some of them is extracted to speed up the planning. The extracted parts include: properties and states of all rooms and furnitures, properties and states of people and objects involved in the goal condition.

In the open world, robots are not omniscient about the state of the world. So one of the problems is that the objects or people in the task are unknown. For instance, in the task \textit{``bring me a pepsi"}, there are no instances of \textit{pepsi}. At this point, from the initial state, the goal is not reachable in the planning. We use \textit{assumptions} to solve the challenge, always assume there are enough objects to meet the goal conditions. We present Algorithm \ref{alg:geninitstate} as an approach to solve this problem by encoding assumptions derived from the goal condition. The algorithm first incorporates instances, attributes, and relations from the current knowledge base in the initial state $ S_0 $ (Line 1-3). Then, for each object referenced by the operator, the require number from operator and the number of known existence are calculated (Line 6-7). The algorithm adds new instances to $ S_0 $ until the number meets the user's requirements (Line 8-18). For each added instance (Line 9), fluents inherited from type (Line 11), goal conditions (Line 13) and assumptions (Line 15) are also added into $ S_0 $.

After the goal and initial state are generated, answer set solver is called to generate answer sets using the union of domain representation, goal conditions and initial fluent set. In the returned answer set, a sequence of actions and fluents that denote expected states before and after execution of actions are obtained.

\begin{algorithm}
    \caption{Initial State Construction}
    \label{alg:geninitstate}
    \begin{algorithmic}[1]
        \REQUIRE Current knowledge $ K $, objects $ O $ from command, goal conditions $ G $, assumptions $ A $
        \ENSURE Initial state $ S_0 $
        \STATE{$ S_0 $ = $ (I,P) $}
        \STATE{$ I $ = \{$ i : i \in K $; $ i $ is an instance\}}
        \STATE{$ P $ = \{$ p : p \in K $; $ p $ is an attribute OR relation\}}
        \FOR{\textbf{each} object name $ o \in O $ }
        \STATE{$ t_o $ = type of $ o $}
        \STATE{$ n $ = number condition of $ o $ in $ G $}
        \STATE{$ m $ = instances' number of $ o_t $ in $ I $}
        \WHILE{$ n > m $}
        \STATE{$ i_o $ = instance with type $ t_o $}
        \STATE{add $ i_o $ to $ I $}
        \STATE{$ P_t $ = relations and attributes of $ t_o $ in $ K $}
        \STATE{add fluents of $ i $ to $ P $ by replacing references to $ t_o $ in $ P_t $}
        \STATE{$ P_o $ = conditions of $ o $ in $ G $}
        \STATE{add conditions of $ i $ to $ P $ by replacing references to $ o $ in $ P_o $}
        \STATE{$ A_o $ = assumptions of $ o $ in $ A $}
        \STATE{add assumptions of $ i $ to $ P $ by replacing references to $ o $ in $ A_o $}
        \STATE{$ n = n - 1 $}
        \ENDWHILE
        \ENDFOR
        \RETURN{$ S_0 $}
    \end{algorithmic}
\end{algorithm}

\subsection{Execution and Monitor}
\begin{figure}[ht]
    \centering
    \includegraphics[width=2.5in]{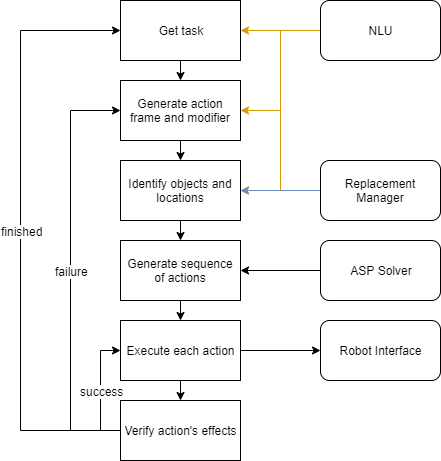}
    \caption{Plan generation, execution, and monitor.}
    \label{fig:plan}
\end{figure}

The components described in the previous sections are employed by the Plan Execution and Monitor component, which is the central coordination component for the command execution. The plan generation and execution follows the traditional ``planning-execution-monitor-replan" loop. A simplified control flow for execution of a planning task is shown in Fig.\ref{fig:plan}.

When a command is sent from a person, NLP module will generate the corresponding  \textit{action frames} and \textit{modifiers}, and then construct the goal conditions. Before planning, parameters in action frames will be confirmed. Combined with the current world state, if the specific entity or location information cannot be confirmed, HRI module will generate query command to the user to confirm these information and rewrite the goal and state based on the confirmed information. Then, domain knowledge, states and goals are transmitted to the planner and the ASP solver generates a plan. If the plan cannot be found, the planner generates a feedback indicating the failure. A successful plan includes a series of actions bound variables, as well as the preconditions and effects of these operations, which will be executed one by one. We use CLINGO \cite{gebserClingoASPControl2014} as our ASP solver.

Each operation corresponding to the symbol planning operator is connected with a state machine that controls the robot's execution. These operations combine elementary or primitive skills into complex skills. The primitive skills are based on services provided by the robot such as inverse kinematics, motion planning, object recognition and location, face recognition, etc. For example,  motion grasping uses a visual servo to accurately perform object localization.

Action execution may fail due to the uncertainty of perception or the change of environment. To illustrate these changes, the preconditions and effects are verified by the \textit{monitor}. The monitoring program detects changes in the environment and the results of action execution, and decide how to make decisions. For environment changes, there are two kinds: one is the effective information for the current task, the other is ineffective to the task. Both will be imported into the knowledge database. We divide the effective information into two kinds, one is the disappearance of preconditions of the actions in the plan, the other is the emergence of new objects or states that meet the requirements of the task. The former represents that the current plan will not be able to complete the task, while the latter indicates that there may be a better solution to complete the task. They will make the robot to replan and try to find a new solution.

If the operation fails due to a missing object, the planner will find a new object to replace the object, for example, another object of the same category or similar. If there are no required objects in the knowledge base, the robot will make the assumptions(assume some required object is somewhere), the same location is only allowed one assumption in a task. In this way, the robot finds the objects that meet the task or reports failure when exit conditions met (timeout, no required objects in any places). Constantly update the state of the world based on sensor data or predictive models. After each operation, the world state observer component is queried to get the current world state. If any mismatch between the planned world state and the current world state is detected, the plan execution is considered to have failed, then re-planning is triggered based on the current world state. If the operation is successful, the next operation will be performed. When the task is completed, the robot enters the idle state and waits for a new command.

\section{Experiments and Results}\label{section:experiments}

To evaluate the system, we have developed a simulation environment with GAZEBO\footnote{\url{http://gazebosim.org}}. The environment contains a domestic house and the model of Kejia. Kejia is equipped with a wheeled mobile base, a single 5-Degree-of-Freedom arm, a 2D laser range finder, an elevator, a pan-tilt, and an RGBD camera. Our system is implemented as nodes in ROS (Robot Operating System). It subscribes all necessary information (arm joint angles, robot pose, recognized objects, etc.) and publishes control messages (navigate to some location, turn on object recognition function to search some object, etc.) which affect the behavior of the robot. The domestic house contains several rooms (bedroom, living room, kitchen, etc.), furniture (bed, table, desk, cupboard, bookshelf, etc.) and some objects (cup, bottle, beer, coke, etc.). For all objects, there are two categories: specific and similar. The specific object has a unique label in the perception system, and this object is bound to this label. Similar objects (apples, same bowl, etc.) share a label in the vision system. For similar objects, their names in the knowledge base are different, the mapping between the vision system and knowledge base depends on location, size, and other properties.

\begin{figure}[ht]
    \centering
    \includegraphics[width=2in]{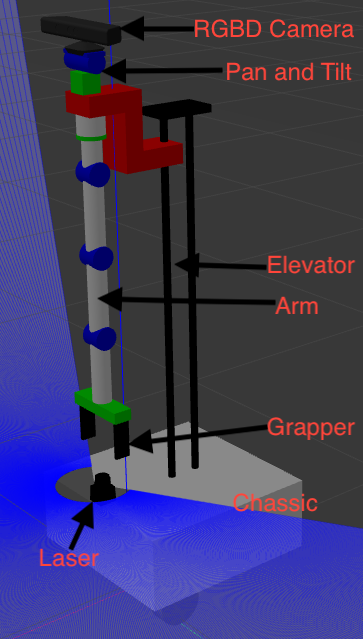}
    \caption{The model of Kejia.}
    \label{fig:robot}
\end{figure}

\begin{figure}[ht]
    \centering
    \includegraphics[width=3.3in]{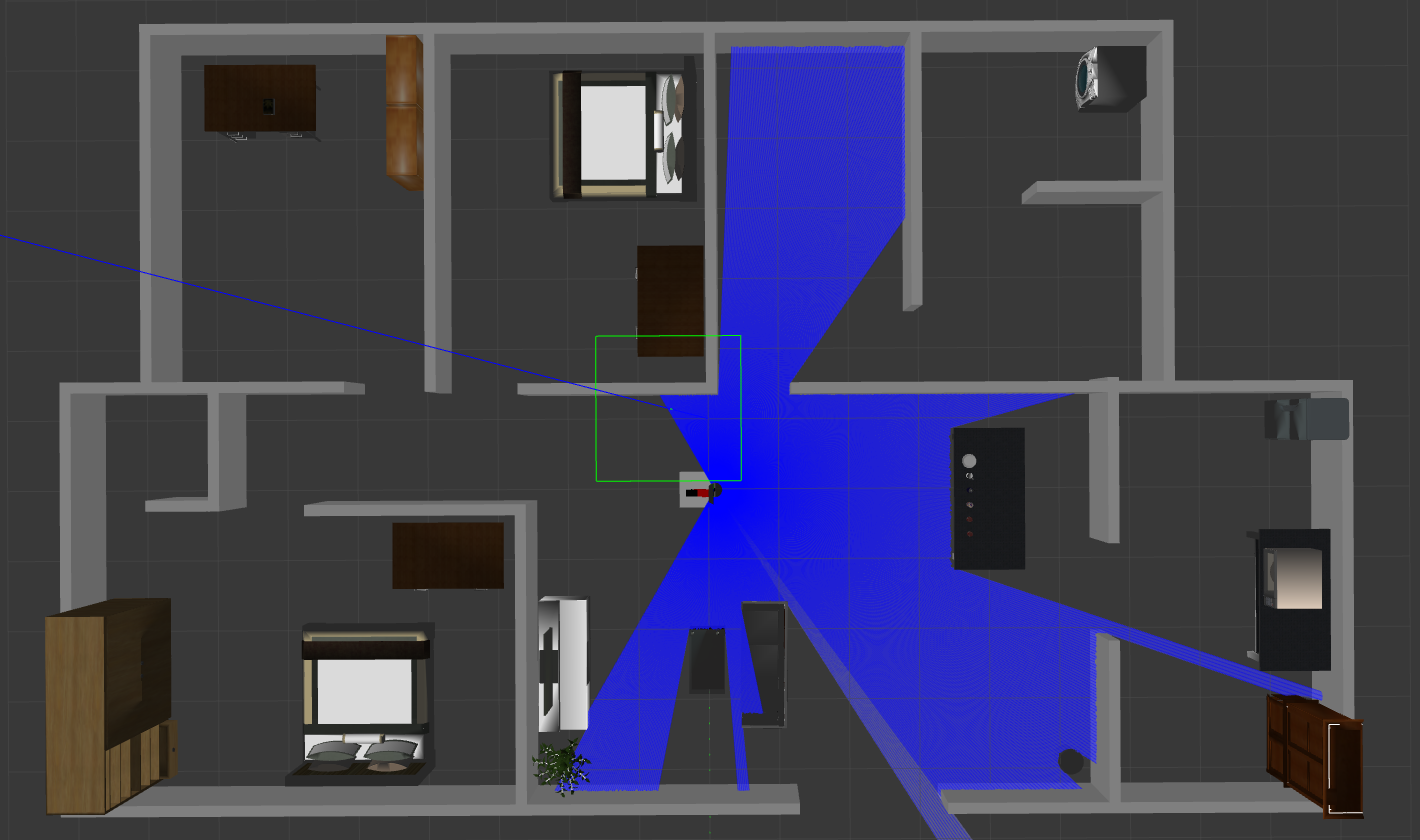}
    \caption{The domestic environment, which includes two bedrooms, one kitchen, one living room, and one study.}
    \label{fig:env}
\end{figure}

In the experiment, we mainly consider three kinds of uncertainties: HRI uncertainties (vague or erroneous information from users), changing environment (object removed from a known position or object appears in an unexpected place), and execution errors (failed grab, or navigation). The required objects and their positions may be unknown to the robot, so the robot needs to assume that there are objects needed in the place given by the user. Here are five scenarios to show how robots respond to these challenges.\footnote{A demo video is available at \url{https://youtu.be/TelJpWJe8q8}}

\noindent\textbf{Scenario 1: Jamie requested a coke (vague information).} Initially Jamie and robot were in the study, the doors from the study to the kitchen are unlocked, the robot knew nothing about coke. The robot got vague information: in the kitchen. The robot first assumed there was coke in the kitchen table, then started searching in the kitchen by visiting the kitchen table first. On the kitchen table, there was one cup, one bottle, and one bowl, but no cokes. The robot made the second assumption: the coke was in the cupboard. It visited the cupboard and found two cokes and one beer, then bring one coke to Jamie.\\
\textbf{Scenario 2: Jamie requested a cup from the living table (error information).} Jamie and the robot were in the study, and Jamie asked the robot to go to the living table and brought a cup. Though the robot knew a cup on the kitchen table, it tried to get a cup from the living table firstly. The robot first assumed there was a cup in the living table, then started searching by visiting the living table. There was nothing on the living table. The robot abandoned the information from Jamie, it had got the information that a cup was on the kitchen table in Scenario 1. It visited the kitchen table and found one cup, one water bottle, and one bowl, then bring the cup to Jamie.\\
\textbf{Scenario 3: Jamie requested a coke (disappearing target).} Jamie and the robot were in the study, the robot knew there was one coke in the cupboard, it visited the cupboard, and started to search the coke, but did not found any coke. The robot removed the item in the knowledge base and made an assumption: there was coke on the kitchen table. However, after it reached the kitchen table, it found no coke on the kitchen table. Then it made another assumption: the coke was on the dining table. The robot navigated to the dining table and found two cokes, finally it taken a coke from the dining table and handed it over to Jamie.
\\
\textbf{Scenario 4: Jamie requested a coke and a beer (unexpected target).} Jamie and the robot were in the study, the robot knew there was one coke in the dining table and a beer in the cupboard, it visited the dining table, and started to search the coke, then it found coke and a beer. An unexpected target (the beer) was found, the robot added a new item to the knowledge base and triggered to replan, a better solution was generated. The robot brought the coke from the dining table to Jamie and then navigated to the dining table, take the beer from the dining table to Jamie.
\\
\textbf{Scenario 5: Jamie requested a bowl (failed grab).} Jamie and the robot were in the study, the robot known there was one bowl in the kitchen table. It visited the kitchen table and started to search the bowl, then it found the bowl. While when the robot picked up the bowl from the kitchen table, it tried twice, but both failed. So it navigated to Jamie and reported this failure.

The demonstration of the robot shows that the system can serve under uncertainties and changing environment by involving assumptions and changing detection. By identifying assumptions or detections that are inconsistent with the knowledge base or not, the robot performs the original plan or makes new assumptions, or replans.

\section{Conclusion}\label{section:conclusion}
In this paper, we present a planning system for a general-purpose service robot, by leveraging HRI, assumptions and continuous sensing, which is aimed at behaving more intelligently in domains that contain incomplete information, under-specified goals, and dynamic changes. Experiments show the robustness of the service robot in the domestic environment. By combining assumption and symbolic planning, the robot can serve without knowing the position of the required object. Proper use of assumptions combined with continuous sensing can be helpful to handle unpredictable domain changes and behavior robust in the open world. In the future, we will address how to make better assumptions to improve planning efficiency.

\bibliographystyle{bib/IEEEtran}
\bibliography{bib/IEEEabrv,bib/my}

\end{document}